\documentclass[sigconf, dvipsnames]{acmart}

\usepackage{tikz}
\usetikzlibrary{shapes, arrows.meta, positioning, shapes.geometric, calc}

\usepackage{multirow}
\usepackage{multicol}
\usepackage{algorithm}
\usepackage{algpseudocode} \usepackage{listings}

\usepackage{cleveref}
\crefformat{section}{#2section~#1#3}
\Crefformat{section}{#2Section~#1#3}
\crefformat{figure}{#2figure~#1#3}
\Crefformat{figure}{#2Figure~#1#3}
\crefformat{table}{#2table~#1#3}
\Crefformat{table}{#2Table~#1#3}
\crefformat{algorithm}{#2algorithm~#1#3}
\Crefformat{algorithm}{#2Algorithm~#1#3}
\crefformat{line}{#2line~#1#3}
\Crefformat{line}{#2Line~#1#3}
\crefformat{appendix}{#2appendix~#1#3}
\Crefformat{appendix}{#2Appendix~#1#3}
\crefformat{equation}{#2equation~#1#3}
\Crefformat{equation}{#2Equation~#1#3}
\crefformat{definition}{#2definition~#1#3}
\Crefformat{definition}{#2Definition~#1#3}
\crefformat{example}{#2example~#1#3}
\Crefformat{example}{#2Example~#1#3}

\AtBeginDocument{}

\copyrightyear{2025}
\acmYear{2025}
\setcopyright{cc}
\setcctype{by-nc-sa}
\acmConference[GRADES-NDA '25]{8th Joint Workshop on Graph Data Management Experiences \& Systems (GRADES) and Network Data Analytics (NDA) }{June 22--27, 2025}{Berlin, Germany}
\acmBooktitle{8th Joint Workshop on Graph Data Management Experiences \& Systems (GRADES) and Network Data Analytics (NDA) (GRADES-NDA '25), June 22--27, 2025, Berlin, Germany}\acmDOI{10.1145/3735546.3735859}
\acmISBN{979-8-4007-1923-3/2025/06}

\begin{document}

\title{Graph Repairs with Large Language Models: An Empirical Study}

\author{Hrishikesh Terdalkar}
\email{hrishikesh.terdalkar@liris.cnrs.fr}
\orcid{0000-0001-8343-424X}
\affiliation{\institution{Lyon1 University, CNRS LIRIS}
  \city{}
  \state{}
  \country{France}
}

\author{Angela Bonifati}
\orcid{0000-0002-9582-869X}
\email{angela.bonifati@liris.cnrs.fr}
\affiliation{\institution{Lyon1 University, CNRS LIRIS \& IUF}
  \city{}
  \country{France}
}

\author{Andrea Mauri}
\orcid{0000-0002-1263-4575}
\email{andrea.mauri@liris.cnrs.fr}
\affiliation{\institution{Lyon1 University, CNRS LIRIS}
  \city{}
  \country{France}
}

\renewcommand{\shortauthors}{Terdalkar et al.}

\begin{abstract}
  Property graphs are widely used in domains such as healthcare, finance, and
  social networks, but they often contain errors due to inconsistencies, missing
  data, or schema violations. Traditional rule-based and heuristic-driven graph
  repair methods are limited in their adaptability as they need to be tailored
  for each dataset. On the other hand, interactive human-in-the-loop approaches
  may become infeasible when dealing with large graphs, as the cost--both in
  terms of time and effort--of involving users becomes too high. Recent
  advancements in Large Language Models (LLMs) present new opportunities for
  automated graph repair by leveraging contextual reasoning and their access to
  real-world knowledge. We evaluate the effectiveness of six open-source LLMs in
  repairing property graphs. We assess repair quality, computational cost, and
  model-specific performance. Our experiments show that LLMs have the potential
  to detect and correct errors, with varying degrees of accuracy and efficiency.
  We discuss the strengths, limitations, and challenges of LLM-driven graph
  repair and outline future research directions for improving scalability and
  interpretability.
\end{abstract}



\received{4 April 2025}
\received[accepted]{2 May 2025}

\graphicspath{{.}{graphics/}}
\maketitle

\section{Introduction}
\label{sec:introduction}

Graphs are powerful data structures for representing relationships between
entities and are widely used in applications such as social networks, healthcare
systems, and knowledge graphs. Among the various graph models, property graphs
have gained significant popularity due to their flexibility in representing
entities as nodes with labeled attributes and relationships as
edges~\cite{2018BFVY,angles2018property}. These graphs form the backbone of many
real-world knowledge
bases~\cite{yamaguchi2014modeling,dai2014using,comyn2017model,tuck2022cancer}.

Despite their utility, property graphs often suffer from errors, including
missing relationships, inconsistent attributes, and structural
violations~\cite{sahu2020ubiquity}. These errors can arise due to data
integration issues, incomplete updates, or inconsistencies in real-world data
sources. Addressing these errors, a process known as \emph{graph repair}, is
crucial for ensuring data integrity and usability. Traditionally, graph repair
relies on application of pre-written rules~\cite{cheng2018rule} and constraint
enforcement~\cite{nebot2012finding,fan2016,fan2017}.

Previous studies have proposed to involve in validating data repairs~\cite{IlyasC19}, with approaches to engage users in the repair process—ranging from knowledge bases~\cite{abdallah,loster21} to labeled graphs~\cite{edbt:igr}, and particularly property graphs~\cite{pachera2025user}. However, these works do not consider that human involvement is expensive both in term of time and cost.

With the rise of Large Language Models (LLMs), new possibilities have emerged
for automated graph repair. LLMs, trained on vast amounts of textual and
structured data, have demonstrated strong reasoning abilities and pattern
recognition
skills~\cite{vaswani2017attention,radford2019language,brown2020language,devlin2018bert,bommasani2021opportunities}.
Recent studies have explored their application in knowledge graph tasks,
including entity resolution and data
augmentation~\cite{yang2023chatgptnotenough,pan2024unifying,yang2024give} and rule mining~\cite{le2025graph}.
However, their effectiveness in repairing property graphs has not been
systematically studied.

\begin{figure}[t]
    \centering
\begin{tikzpicture}[
        publicnode/.style={circle, rounded corners, draw=blue, thick, minimum size=5mm},
        privatenode/.style={circle,rounded corners, draw=blue, thick, minimum size=5mm},
        publicedge/.style={draw=blue, thick, ->, >=Stealth},
        privateedge/.style={draw=blue, thick, ->, >=Stealth},
        erroredge/.style={draw=Red, ultra thick, ->, >=Stealth},
        ]

\node[publicnode]  (disease)       {D1};
        \node[publicnode]  (medicine)      [right=of disease, xshift=1.25cm,] {M1};
        \node[publicnode]  (medicine-2)      [left=of disease, xshift=-1cm] {M2};
        \node[privatenode]  (patient)      [below=of disease, yshift=-0.5cm] {P1};
        \node[publicnode]  (ingredient)    [below=of medicine, yshift=-0.5cm] {I1};
        \node[publicnode]  (ingredient-2)    [below=of medicine-2, yshift=-0.5cm] {I2};

\draw[erroredge] (patient) -- node[midway, sloped, fill=white] {takes} (medicine);
        \draw[erroredge] (patient) -- node[midway, sloped, fill=white] {allergic-to} (ingredient);
        \draw[erroredge] (medicine) -- node[midway, fill=white] {contains} (ingredient);
        \draw[publicedge] (medicine-2) -- node[midway, fill=white] {contains} (ingredient-2);
        \draw[privateedge] (patient) -- node[midway, fill=white] {suffers-from} (disease);
        \draw[publicedge] (medicine) -- node[midway, sloped, fill=white] {treats} (disease);
        \draw[publicedge] (medicine-2) -- node[midway, sloped, fill=white] {treats} (disease);

\node (patient-label) at (patient.south) [below] {\footnotesize :Patient};
        \node at (ingredient.south) [below] {\footnotesize :Ingredient};
        \node at (ingredient-2.south) [below] {\footnotesize :Ingredient};
        \node at (disease.north) [above] {\footnotesize :Disease};
        \node (medicine-label) at (medicine.north) [above] {\footnotesize :Medicine};
        \node at (medicine-2.north) [above] {\footnotesize :Medicine};

\draw[thick,dotted] ($(disease.north west)+(-0.6,0.6)$) rectangle ($(ingredient.south east)+(0.5,-0.6)$);
\node at ($(disease.north east)+(0.5,0.6)$) [above right] {\footnotesize Inconsistency ($i_j$)};

    \end{tikzpicture}
    \caption{Inconsistency in a Property Graph: A patient is being given
    medicine containing an ingredient to which the patient is allergic.
    Potential sources of error include the allergy edge, the ingredient edge or
    more. The inconsistency may be repaired in several ways, including but not
    limited to, the removal of one of the three edges in red.}
    \label{fig:inconsistency-example}
\end{figure}
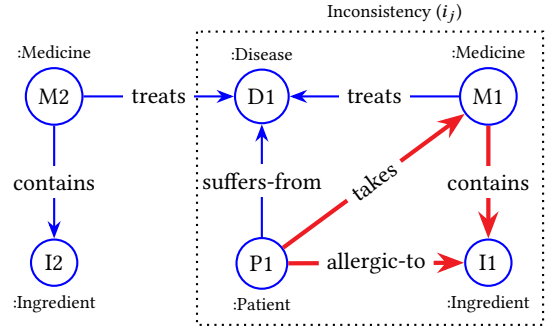

\begin{example}
Consider a knowledge graph representing patients, diseases, medications, and
allergies. Each vertex represents an entity, and each edge represents a
relationship. \Cref{fig:inconsistency-example} illustrates a scenario where
inconsistencies arise in such a graph. Specifically, a patient is prescribed a
medication that contains an ingredient they are allergic to. Such
inconsistencies can stem from various sources of error, including
(i)~\emph{Incorrect ingredient information:} The medication's composition may be
inaccurately recorded, or (ii)~\emph{Incorrect allergy information:} The
patient's allergy data might be erroneous.
Ideally LLMs, with their broad exposure to biomedical and pharmaceutical
knowledge, can assist in identifying and rectifying these errors. Given a
structured query about the medication's ingredients, an LLM can cross-verify
this information against its inherent knowledge. If the recorded ingredient is
incorrect, the LLM can suggest a correction. Similarly, while patient-specific
allergies require external validation, LLMs can still infer potential
inconsistencies through reasoning.
\end{example}

\begin{figure*}[t]
    \includegraphics[width=\linewidth]{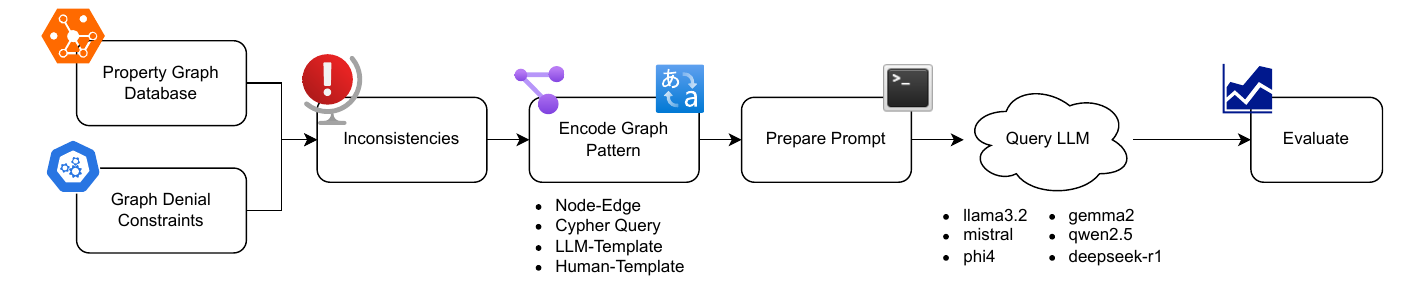}
    \caption{Repairing Property Graph Inconsistencies with Large Language Models}
    \label{fig:overview}
\end{figure*}

In this work, we evaluate the capabilities of LLMs to repair property graphs. In particular, we focus on Graph Denial Constraints~\cite{fan2017} specifying a condition or situation that must not happen, e.g. a patient must not
be prescribed a medicine containing an ingredient to which they are allergic.

We experiment with multiple LLMs, including LLaMA~\cite{llama3_meta},
Mistral~\cite{mistral_ai}, Phi~\cite{phi4_microsoft},
Gemma~\cite{gemma2_google}, Qwen~\cite{qwen2_alibaba}, and
DeepSeek~\cite{deepseek_r1} to determine their strengths and limitations in
graph repair tasks. Our evaluation uses a modified version of a synthetic
medical dataset, enriched with medical ingredient data and allergy information,
to simulate real-world inconsistencies.
We experiment with different graph encodings, from simple incident encoding~\cite{talklikeagraph}, human readable template, to an agent-based encoding where LLMs generate a description of the graph.
We evaluate the performances in terms of the quality of the repair, adherence to the
output format, runtimes and the size of the output.

Our findings highlight that each LLM responds differently to the prompt setup.
Overall, with a correct choice, models are able to adhere to the instructed
output format with high success (ranging between $72\%$ to $96\%$) and produce
`valid' repairs that would get rid of the inconsistency with moderate success
($42\%$-$74\%$), there is still a long way to reach a point where LLMs can
autonomously perform `accurate' graph repairs. Among the chosen LLMs,
\texttt{deepseek-r1} showed the best results, however with the trade-off of
considerable computational cost. Among others, \texttt{phi4} and \texttt{gemma2}
show potential to produce comparable results under the appropriate choices of
graph encoding method and output specification.

The remainder of this paper is organized as follows: \Cref{sec:related} provides
an overview of related work on graph repair and the application of LLMs.
\Cref{sec:pipeline} details the architecture of our proposed pipeline.
\Cref{sec:experiments} outlines the dataset, experimental setup, and evaluation
methodology. The experimental results are presented in \Cref{sec:results},
followed by an in-depth discussion and directions for future research in
\Cref{sec:discussion}. Finally, \Cref{sec:conclusion} summarizes the key
findings, and \Cref{sec:limitations} discusses the study's limitations. 
\section{Related Work}
\label{sec:related}

This section reviews the existing literature on graph repair and the application
of Large Language Models in graph related tasks.

\vspace*{2mm}
\noindent{\textbf{Graph Repair.}} Due to the ubiquity of errors in graphs,
various techniques for error detection and correction have gained
importance~\cite{sahu2017ubiquity}. Traditional approaches to graph repair often
rely on rule-based systems and constraint enforcement
\cite{cheng2018rule,song2017graph}.
Human-in-the-loop methods, integrating human insight in combination with
rule-based methods are useful when specific domain-knowledge requirement
restricts the applicability of comprehensive rule-based approaches
\cite{oosterman2014crowdsourcing,samiotis2020microtask}, or if the existing
algorithms need improvement \cite{bhardwaj2022human}. Graph Denial Constraints
play a crucial role in specifying integrity constraints for graphs. There has
been significant work on the theory and application of denial constraints in
databases and knowledge graphs \cite{bertossi2011database}. In recent times,
interactive methods to repair neighborhood constraints
\cite{juillard2024interactive}, and denial constraints \cite{pachera2025user}
have been proposed.
Our work differs from these traditional approaches by leveraging the reasoning
capabilities of LLMs to infer and correct inconsistencies, rather than relying
solely on predefined rules.

\vspace*{2mm}
\noindent{\textbf{Large Language Models and Knowledge Graphs.}} Integrating LLMs with graph
processing has attracted considerable research interest. A recent survey~\cite{2024survey}
highlights promising future research directions, including the development of
multi-modal LLMs capable of processing diverse data types, which could
significantly improve reasoning capabilities over graphs. LLMs have been shown
to be able to understand graph-structured information \cite{cheng2018rule}. Use of
natural language to interact with graphs has also shown promising results
\cite{peng2024chatgraph}. Also, recent studies highlighted the capabilities of LLMs to extract logical and constraints from graph structure~\cite{le2025graph,luo2023chatrule}
These studies highlight the growing potential of LLMs in graph-related tasks,
while also acknowledging the existing challenges. Our work - to the best of our knowledge - is the first that focuses on the
specific task of graph repair, leveraging the inherent `real-world knowledge'
and reasoning capabilities of LLMs to address semantic inconsistencies in
property graphs.

\vspace*{2mm}
\noindent{\textbf{Prompt Engineering for Structured Data.}} Prompt engineering is an emerging discipline concerned with crafting and
fine-tuning prompts to unlock the full potential of LLMs in diverse applications
and research areas. Prompt engineering has utility for several tasks in
specialized domains including healthcare
\cite{mesko2023prompt,wang2023prompt,venerito2024prompt}, chemistry
\cite{hatakeyama2023prompt}, and several others
\cite{wu2023exploring,giray2023prompt,marvin2023prompt}. Prompt engineering can
also serve as a tool for utilizing LLMs in combination with knowledge graphs to
perform analytical tasks \cite{jiang2023structgpt,fatemi2023talk}. Encoding
labelled graphs as text as a subroutine of prompt-engineering in order to
produce better responses has become a popular area of work
\cite{jiang2023structgpt,fatemi2023talk}.
Our approach builds on these insights by designing prompts that effectively
convey the graph structure and repair requirements to LLMs. We also explore the
impact of few-shot learning by providing examples of the desired output format.

\section{Large Language Models for Graph Repair}
\label{sec:pipeline}

In this work, we focus our attention on the repairs in knowledge graphs
following the property graph data model.

A \emph{property graph}~\cite{angles2018property} is a directed, labeled, and attributed multi-graph where
both vertices and edges are objects that carry labels and key-value attributes.
Formally, a property graph is defined as $G = (V, E, \eta, \lambda, \nu)$,
where: $V$ and $E$ are disjoint sets of vertex and edge objects, respectively,
$\eta: E \rightarrow V \times V$ maps each edge to an ordered pair of vertices,
$\lambda: V \cup E \rightarrow \mathcal{P}(\mathcal{L})$ assigns a set of
labels, and $\nu: (V \cup E) \times \mathcal{K} \rightarrow \mathcal{N}$ assigns
key-value properties.

A property graph $G$ is said to be \emph{inconsistent} if it violates a given
set of constraints $\Psi$. These constraints specify logical statements that
must hold over the graph structure and attributes.

A common type of constraint is a \emph{Graph Denial Constraint (GDC)}~\cite{fan2017},
expressed as $\phi = (Q[\Bar{x}], X \rightarrow Y)$, where, $Q[\Bar{x}]$ is a
graph pattern describing the topology of an inconsistency, $X$ and $Y$ are sets
of literals that impose attribute conditions. An inconsistency occurs when $X$
holds, but $Y$ does not, violating the constraint.

Given a property graph $G$ and a set of constraints $\Psi$, a \emph{graph
repair} is a sequence of update operations that transform $G$ into a new graph
$G'$ such that $G'$ satisfies $\Psi$. Update operations can include
modifications such as addition or deletions of nodes or edges, label changes,
and property updates.

\subsection{Repair Process}

\Cref{fig:overview} illustrates the overall repair process. A
denial constraint is described in the form of a \emph{Cypher} query
\cite{cypher,hua2023gdsmith} as in~\cite{pachera2025user}. Thus, every match found for the query represents a
violation of the constraint. We then encode the subgraph corresponding to the violation in a text format to be
provided to LLMs. An LLM is provided (i)~system prompt clarifying the task,
(ii)~description of the inconsistency, and (iii)~description of the output
format. The responses from LLMs are then evaluated in terms of the validity, the
accuracy and the cost of repairs (e.g, \#tokens, \#operations, and evaluation time).

\subsubsection{Inconsistency Detection}

We employ Cypher queries to identify instances of GDC violations which model the
semantic inconsistencies in property graphs. For example, the following query
detects patient-medication inconsistencies:

\begin{lstlisting}
MATCH (p:Patient)-[rm:TAKES]->(m:Medication),
        (m)-[rc:CONTAINS]->(i:Ingredient),
        (p)-[ra:ALLERGIC_TO]->(i)
RETURN *
\end{lstlisting}

\subsubsection{Prompt Engineering}

We carefully structure the prompts to ensure consistent LLM-generated repairs.
The prompt comprises the following components: (i)~\emph{system prompt} that
instructs the LLM to utilize real-world knowledge to repair inconsistencies
(ii)~\emph{encoding of the inconsistency} in natural language, and
(iii)~\emph{description of the output format}

\paragraph{\textbf{System Prompt}}

The system prompt instructs the LLM on its role and the task at hand,
emphasizing the use of real-world knowledge for accurate repairs. Minor
variations of the following system prompt are used, with
\texttt{INPUT\_DESCRPITION} describing the input encoding format.

\begin{lstlisting}
You are an AI assistant for Graph Repair. Your
task is to identify factual inconsistencies and
suggest corrections using structured graph
operations.

You will receive:
{INPUT_DESCRIPTION}

Allowed repair operations:
1. `ADD_NODE` - Add a new node
2. `DEL_NODE` - Remove a node
3. `ADD_EDGE` - Add a new edge between nodes
4. `DEL_EDGE` - Remove an existing edge
5. `UPD_NODE` - Modify a node's properties
6. `UPD_EDGE` - Modify an edge's properties

Suggest only factually accurate repairs. Use the
provided format for output. Keep the suggested
number of repair operations small.
\end{lstlisting}

\paragraph{\textbf{Encoding the Inconsistency}}
\label{sec:encoding-modes}

We explore three methods to encode the graph for LLM input:
(i)~\emph{M1: Node-Edge Representation}: a direct node-edge representation of
the subgraph corresponding to the inconsistency~\cite{talklikeagraph}, (ii)~\emph{M2: Template-Based
Text Description}: a human-readable description of the inconsistency generated
using predefined templates, and (iii)~\emph{M3: LLM-Generated Description}:
Natural language descriptions generated by LLMs from the node-edge
representation

In the first method, a graph is represented as a list of lines with each line
representing a node or an edge. The line begins with the keywords \texttt{Node}
or \texttt{Edge}, followed by comma-separated key-value pairs denoting the
properties of the object. The following is an example of M1 encoding.

\begin{lstlisting}
Node i:6700 labels: frozenset({'Ingredient'}), properties: {'id': 'verapamil'}
Node m:6699 labels: frozenset({'Medication'}), properties: {'description': 'verapamil hydrochloride 40 MG Oral Tablet', ...}
Node p:6588 labels: frozenset({'Patient'}), properties: {'first': 'Rosio404', 'last': 'Bayer639', ...}
Edge rc:6699 -> 6700  type: HAS_INGREDIENT {...}
Edge ra:6588 -> 6700  type: ALLERGIC_TO {...}
Edge rm:6588 -> 6699  type: TAKES_MEDICATION {...}
\end{lstlisting}

In the second method, a predefined template is used which describes the
inconsistency along with the variables used to refer to the entities from a
particular match. An example of such an encoding appears below.

\begin{lstlisting}
A person should not be treated with a medicine
that contains an ingredient that the person is
allergic to. However, a person (p)
(p.first=Sanford861) takes a medicine (m)
(m.description=1 ML Epogen 4000 UNT/ML Injection)
which contains an ingredient (i) (i.id=oxycodone)
and (p) is allergic to (i).
\end{lstlisting}

In the third method, we provide the node-edge representation (M1) of an
inconsistency to an LLM and instruct it to encode the inconsistency as natural
language text. For completeness, we evaluate the performance by encoding each
inconsistency using all 6 LLMs.

\paragraph{\textbf{Description of the Inconsistency}}

We evaluate \textbf{four} prompting setups, varying the information provided to the LLM.
The LLM-aided encoding (M3) actually corresponds to 6 more setups, thereby resulting in
\textbf{nine} total prompting setups.

\begin{enumerate}
    \item Only M1
    \item Cypher query and matched result
    \item Cypher query + M2
    \item[(4-9)] Cypher query + M3 (x6)
\end{enumerate}

\paragraph{\textbf{Output Format}}

We define a custom output format, prompting an LLM to provide repairs in the
following structured format.

\begin{lstlisting}
<repairs>
{op_code} | {target} | {details}
</repairs>
\end{lstlisting}

The type of repair operation (\texttt{op\_code}) may be one of the six
options--`ADD\_NODE`, `DEL\_NODE', `ADD\_EDGE`, `DEL\_EDGE`, `UPD\_NODE`, and
`UPD\_EDGE`. The \texttt{target} variable specifies the affected node or
relationship variable referenced in the inconsistency description and
\texttt{details} field contains the relevant property changes in a
\texttt{key=value} format. The instruction to generate repairs within the
`\texttt{<repairs>}' tag ensures that even if an LLM generates more text than
necessary, it is easier to parse the response in a systematic manner.

\subsubsection{Few Shots at Example Output Format}
\label{sec:example-modes}

To test the effect of provided examples, we use \textbf{zero-shot} and \textbf{few-shot} prompting techniques
providing zero, one or two examples of output format. We experiment with \textbf{five} example modes.

\begin{itemize}
\item \textbf{none}: \textbf{No} examples
\item \textbf{1-small}: \textbf{One} example of potential repairs containing a single repair operation
\item \textbf{2-small}: \textbf{Two} examples of potential repairs containing a single repair operation each
\item \textbf{1-large}: \textbf{One} examples of potential repairs containing multiple repair operations
\item \textbf{2-mixed}: \textbf{Two} examples--one small and one large--of potential repairs
\end{itemize}
 
\section{Experiments}
\label{sec:experiments}

The experimental setup consists of a pipeline where inconsistencies in property
graphs are identified using denial constraints, the graph is represented in a
text form using multiple encoding methods. The results are evaluated across six
different LLMs to determine capabilities and trends using various metrics.

\subsection{Dataset}

\emph{Synthea} is a tool designed to simulate realistic patient populations,
generating comprehensive medical records that mirror the demographic and health
characteristics of real-world populations, exemplified by its
\emph{SyntheticMass} dataset \cite{walonoski2018synthea}. This synthetic data
encompasses a wide range of patient life events, incorporating details such as
diseases, healthcare encounters, medication use, allergies and more. It produces
data in standard healthcare formats, facilitating research and analysis without
compromising patient privacy. The codes and descriptions of medications are
referenced from the \emph{RxNorm} database.

\emph{RxNorm}, serves as a standardized nomenclature for medications,
establishing a consistent naming system for both generic and brand-name drugs
\cite{liu2005rxnorm}. RxNorm aggregates drug information from various sources,
assigning unique identifiers and provides information about drugs such as their
active ingredients, brands and more.

To introduce inconsistencies, we modify the \emph{Synthea} dataset by adding
errors related to ingredient allergies. We refer to this modified dataset as
\emph{Synthea++} hereafter. We first obtain correct ingredients for every
medication from RxNorm database, and then with a controlled probability
($p_{wrong\_ingredient} = 0.15$) add incorrect ingredients. Further, we
artificially introduce allergies to these ingredients in the following manner.
For every entry related to a medicine consumption, we define two parameters,
$p_{allergy} = 0.05$ as the probability for adding an allergy to ingredients of
the prescribed medication and $p_{wrong\_allergy} = 0.25$ as the probability to
introduce allergy to an incorrect ingredient. Controlled by these parameters, we
introduce $\sim 10\%$ (total $165$) inconsistencies.

\begin{table}
    \caption{Dataset Statistics}
    \label{table:dataset}
\begin{tabular}{llc}
\toprule
        \bf Label Type & \bf Label & \bf \#Instances\\
        \midrule
        \multirow{4}{*}{Node Labels} & Patient & $1171$ \\
        ~ & Medication & $131$ \\
        ~ & Ingredient & $113$ \\
        ~ & Allergy & $15$ \\
        \midrule
        ~ & \bf Total & $1430$\\
        \midrule
        \multirow{4}{*}{Edge Labels} & TAKE\_MEDICATION & $1000$\\
        ~ & ALLERGIC\_TO & $27$ \\
        ~ & HAS\_INGREDIENT & $195$\\
        \midrule
        ~ & \bf Total & $1222$\\
        \bottomrule
    \end{tabular}
\end{table}

\paragraph{\textbf{Correct Repair}}

The constraint we impose is that a patient should not be treated with a
medication that contains an ingredient that they are allergic to. For every
violation of such a constraint, the ingredient that the patient is allergic to
is either a real ingredient or an incorrect one introduced by us. In the former
case, the allergy is treated as incorrect, and the removal of the allergy-edge
is considered to be the `correct' repair. While in the latter case, the removal
of the ingredient-edge is the `correct' repair.

\subsection{Models}

We evaluate six open-source LLMs for their ability to generate
repair suggestions. The selected models are:

\begin{enumerate}
    \setlength{\leftskip}{-15pt}       

    \item \textbf{LLaMA 3.2} (Meta): A general-purpose, open-weight AI model with strong fine-tuning capabilities, supporting multimodal tasks and optimized for efficiency \cite{llama3_meta}.
    \item \textbf{Mistral}: A lightweight yet powerful model designed for efficient instruction-following, excelling in task-specific adaptability and performance \cite{mistral_ai}.
    \item \textbf{Phi 4} (Microsoft): Optimized for structured reasoning tasks, including problem-solving and mathematical reasoning, with a strong focus on interpretability and reliability \cite{phi4_microsoft}.
    \item \textbf{Gemma 2} (Google): Specializes in structured data analysis and knowledge graph understanding, making it well-suited for enterprise and research applications \cite{gemma2_google}.
    \item \textbf{Qwen 2.5} (Alibaba): A multilingual AI model designed for handling structured data and cross-lingual applications, optimized for diverse linguistic tasks \cite{qwen2_alibaba}.
    \item \textbf{DeepSeek R1}: Focuses on domain-specific reasoning with an emphasis on logical consistency, making it suitable for research and specialized industry use cases \cite{deepseek_r1}.

\end{enumerate}

\begin{table}
    \caption{Details of LLMs}
    \label{table:modeldetail}
    \begin{tabular}{lccc}
        \toprule
        \bf Model   & \bf Ollama ID & \bf Parameters & \bf Size\\
        \midrule
        llama3.2    & \tt a80c4f17acd5  &  7B            & 2.0 GB  \\
        mistral     & \tt f974a74358d6  &  7B            & 4.1 GB  \\
        phi4        & \tt ac896e5b8b34  &  14B           & 9.1 GB  \\
        gemma2      & \tt ff02c3702f32  &  9B            & 5.4 GB  \\
        qwen2.5     & \tt 845dbda0ea48  &  7B            & 4.7 GB  \\
        deepseek-r1 & \tt 0a8c26691023  &  7B            & 4.7 GB  \\
        \bottomrule
    \end{tabular}
\end{table}

We use \emph{Ollama 0.6.2} to run LLMs on \emph{MacBook Pro M3 Max} running
\emph{MacOS Sonoma 14.5} with 36GB memory and $14$ GPU cores. In each case, we
choose the latest default model of the corresponding model. Each prompt to the
LLM is made with the \emph{temperature = 0.4} to ensure more consistent
responses. \Cref{table:modeldetail} provides details of the models including
their \emph{Ollama ID}, number of parameters and size.

The graph is stored in a \emph{Neo4j Graph Database}
(neo4j-community-5.23.0)~\cite{webber2012neo4j} and queried using  \emph{neo4j
5.24.0} Bolt driver with \emph{Python 3.11.5}. The models are prompted using the
Python library \emph{langchain-ollama 0.2.4}. The source code is available
here~\cite{llm-graph-repair}.

\subsection{Evaluation Metrics}
\label{sec:evaluation}

The evaluation is structured around three primary objectives: consistency of
formatting, correctness of repairs,  and efficiency of repair generation. Each
model is evaluated based on \textbf{adherence} to the instructed format (denoted
by $F$), \textbf{validity} of suggested repairs (denoted by $V$),
\textbf{accuracy} of suggested repairs (denoted by $A$), and \textbf{cost} in
terms of \#tokens, \#operations, and evaluation time. We also observe trends in
the distribution of suggested types of repair operations and draw
insights in correlation to the accuracy.

\subsubsection{Repair Validity and Correctness}

A response is considered to be in the \emph{valid format} ($F = 1$) if it
adheres to the expected output format using the required structured syntax. The
\emph{validity} and \emph{correctness} of a repair is assessed by comparing it
against predefined ground-truth corrections. A repair is termed to be
\emph{valid} ($V = 1$) if the suggested repairs result in the elimination of the
violation of the denial constraint, even if the repair might not match the
ground-truth. A repair is termed to be \emph{correct} ($A = 1$) if it exactly
matches the ground-truth repair. The binary scores for these criteria are
averaged over all inconsistencies.

\subsubsection{Computational Cost}

The computational cost of generating the repairs is evaluated in terms of the
time, and the token usage. The computational cost of generating the repairs is
evaluated in terms of processing time and token usage. Specifically, we
compare--(i)~\emph{prompt evaluation time} (the time taken to process the input
prompt), (ii)~\emph{evaluation time} (the time taken to generate repairs),
(iii)~\emph{token count} in the generated response, and (iv)~\emph{number of
repair operations} suggested in the response. The time is reported in seconds to
repair one inconsistency, averaged over all inconsistencies.
 
\section{Results}
\label{sec:results}

\begin{table*}[h]
    \centering
    \caption{Cost of LLMs Averaged Across 15 Runs Varying Encoding Modes and Few-Shot Example Modes}
    \label{table:llm_cost}
\begin{tabular}{lcccccc}
        \toprule
        \bf Model & \bf Prompt Time & \bf Evaluation Time & \bf \#Tokens & \bf \#Characters & \bf \#Characters & \bf \#Repair Operations\\
        ~ & (sec/prompt) & (sec/prompt) & ~ & (Response) & (Repair) & ~\\
        \midrule
        llama3.2 & 0.24 & 2.09 & 151.0 & 383.0 & 354.9 & 8.7\\
        mistral & 0.63 & 3.87 & 182.9 & 480.4 & 236.6 & 4.6\\
        phi4 & 1.07 & 6.23 & 134.2 & 618.2 & 57.9 & 1.9\\
        gemma2 & 0.73 & 1.26 & 44.4 & 82.3 & 57.5 & 1.1\\
        qwen2.5 & 0.56 & 1.00 & 42.6 & 95.4 & 71.3 & 1.6\\
        deepseek-r1 & 0.57 & 11.40 & 478.6 & 2107.9 & 53.5 & 1.4\\
        \bottomrule
    \end{tabular}
\end{table*}

\begin{figure}
    \includegraphics[width=\columnwidth]{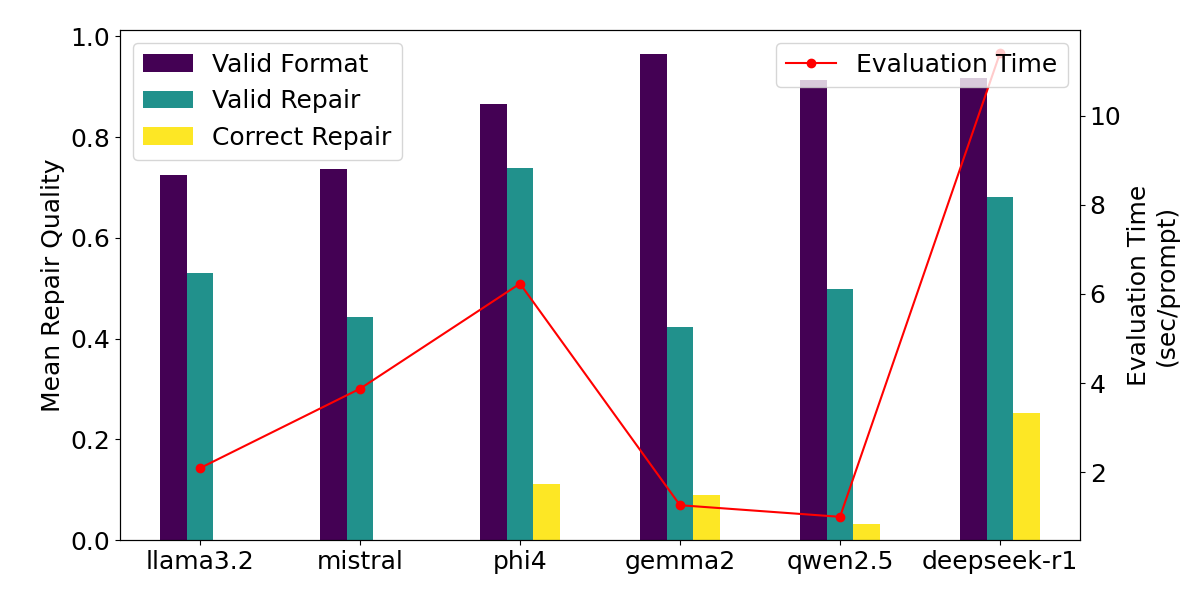}
    \caption{Repair Quality of LLMs Averaged Across 15 Runs Varying Encoding Modes and Few-Shot Example Modes}
    \label{fig:llm_comparison}
\end{figure}

We evaluate the LLMs across multiple runs by varying prompt setup (i.e.,
encoding modes and few-shot examples). \Cref{table:llm_cost} reports the overall
cost of each model--in terms of generation time and response length--averaged
across these runs. \Cref{fig:llm_comparison} illustrates the repair quality of
each model averaged across the same executions. Several models are consistently
($>90\%$ of the times) able to adhere to the required format. While the accuracy
still needs a significant improvement, we observe that \texttt{deepseek-r1} is
fairly resistant to variations in prompt setups and few-shot examples. This,
however, comes at a significant cost in terms of time and tokens generated.

\subsection{Ablation with Encoding Modes and Few-Shot Examples}

\Cref{table:encoding_ablation} presents a comparison of different encoding modes
(\Cref{sec:encoding-modes}) and their impact on evaluation metrics, including
evaluation time, adherence to output format, and validity and accuracy of
repairs. \Cref{table:example_ablation} explores the effect of varying the number
and type of examples (\Cref{sec:example-modes}).

Overall, template-based encoding modes tend to yield higher validity and
correctness scores, especially when combined with a few examples. The Cypher
encoding mode also demonstrates better performance in certain cases,
highlighting the ability of LLMs to understand graph descriptions. It should be
noted however that encoding a graph with LLMs incurs more computational cost.
\Cref{table:llm_encoding_evaluation} illustrates the statistics of encoding a
graph inconsistency using LLMs in terms of length of the encoded prompt and time
taken for encoding.

The results with few-shot example outputs indicate that while a larger number of
examples can sometimes enhance performance, diminishing returns or even
performance degradation can occur if examples are not optimally selected.

\begin{table*}[h]
    \centering
    \caption{Comparison of Quality across LLMs and Encoding Modes (Quality is measured using the adherence to the output format~(F), the validity of the repair~(V), and the accuracy of the repair~(A))}
    \label{table:encoding_ablation}
    \resizebox{\linewidth}{!}{
    \begin{tabular}{l|ccc|ccc|ccc|ccc|ccc|ccc}
        \toprule
        \bf Encoding Mode & \multicolumn{3}{c}{\bf llama3.2} & \multicolumn{3}{c}{\bf mistral} & \multicolumn{3}{c}{\bf phi4} & \multicolumn{3}{c}{\bf gemma2} & \multicolumn{3}{c}{\bf qwen2.5} & \multicolumn{3}{c}{\bf deepseek-r1}\\
        ~ & \bf F & \bf V & \bf A & \bf F & \bf V & \bf A & \bf F & \bf V & \bf A & \bf F & \bf V & \bf A & \bf F & \bf V & \bf A & \bf F & \bf V & \bf A\\
        \midrule
        Graph & 0.79 & 0.51 & 0.00 & 0.67 & 0.02 & 0.00 & \bf 0.91 & 0.56 & 0.18 & \bf 1.00 & 0.00 & 0.00 & \bf 1.00 & 0.88 & 0.00 & \bf 0.95 & 0.62 & 0.24\\
        Cypher & \bf 0.93 & 0.58 & 0.00 & 0.61 & 0.24 & 0.00 & 0.80 & 0.77 & \bf 0.36 & \bf 1.00 & 0.00 & 0.00 & \bf 0.99 & 0.02 & 0.00 & \bf 0.95 & 0.76 & \bf 0.32\\
        Template & 0.74 & 0.68 & 0.00 & \bf 0.98 & \bf 0.98 & 0.00 & 0.83 & 0.83 & 0.00 & \bf 1.00 & \bf 1.00 & 0.00 & \bf 0.96 & \bf 0.91 & 0.00 & \bf 0.93 & 0.84 & \bf 0.38\\
        \midrule
        LLM (llama3.2) & 0.72 & 0.46 & 0.01 & 0.58 & 0.20 & 0.00 & 0.75 & 0.55 & 0.12 & \bf 0.98 & 0.19 & 0.05 & \bf 0.93 & 0.19 & 0.01 & \bf 0.91 & 0.68 & 0.23\\
        LLM (mistral) & 0.71 & 0.50 & 0.00 & 0.45 & 0.21 & 0.00 & 0.77 & 0.58 & 0.16 & \bf 0.99 & 0.09 & 0.06 & \bf 0.98 & 0.28 & 0.01 & \bf 0.91 & 0.48 & 0.10\\
        LLM (phi4) & 0.58 & 0.36 & 0.00 & 0.58 & 0.25 & 0.00 & 0.90 & 0.70 & \bf 0.37 & \bf 1.00 & 0.22 & 0.16 & \bf 0.97 & 0.07 & 0.02 & \bf 0.94 & 0.63 & 0.25\\
        LLM (gemma2) & 0.68 & 0.50 & 0.00 & 0.70 & 0.62 & 0.00 & 0.87 & 0.87 & 0.04 & \bf 1.00 & 0.58 & 0.14 & \bf 0.95 & 0.50 & 0.03 & \bf 0.92 & 0.83 & \bf 0.32\\
        LLM (qwen2.5) & 0.76 & 0.44 & 0.00 & 0.55 & 0.20 & 0.00 & 0.83 & 0.78 & 0.15 & \bf 0.93 & 0.07 & 0.05 & \bf 0.95 & 0.08 & 0.04 & \bf 0.95 & 0.74 & \bf 0.34\\
        LLM (deepseek-r1) & 0.68 & 0.44 & 0.00 & 0.62 & 0.13 & 0.00 & 0.78 & 0.70 & 0.18 & \bf 0.98 & 0.12 & 0.02 & \bf 0.94 & 0.12 & 0.03 & 0.90 & 0.70 & 0.25\\
        \bottomrule
    \end{tabular}
    }
\end{table*}

\begin{table}[h]
    \centering
    \caption{Graph Encoding Performance across LLMs}
    \label{table:llm_encoding_evaluation}
    \resizebox{\columnwidth}{!}{
    \begin{tabular}{lcccc}
        \toprule
        \bf Model & \bf \#Tokens & \bf \#Words & \bf \#Lines & \bf Encoding Time\\
        ~ & ~ & ~ & ~ & (sec/prompt)\\
        \midrule
        llama3.2 & 151.0 & 54.8 & 10.9 & 2.09\\
        mistral & 182.9 & 62.2 & 8.0 & 3.87\\
        phi4 & 134.2 & 90.5 & 10.2 & 6.23\\
        gemma2 & 44.4 & 9.3 & 5.6 & 1.26\\
        qwen2.5 & 42.6 & 12.5 & 3.5 & 1.00\\
        deepseek-r1 & 478.6 & 336.9 & 24.6 & 11.40\\
        \bottomrule
    \end{tabular}
    }
\end{table}

\begin{table*}[h]
    \centering
    \caption{Comparison of Quality across LLMs and Few-shot Example Output Prompts (Quality is measured using the adherence to the output format~(F), the validity of the repair~(V), and the accuracy of the repair~(A))}
    \label{table:example_ablation}
    \resizebox{\linewidth}{!}{
    \begin{tabular}{l|ccc|ccc|ccc|ccc|ccc|ccc}
        \toprule
        \bf \#Examples & \multicolumn{3}{c}{\bf llama3.2} & \multicolumn{3}{c}{\bf mistral} & \multicolumn{3}{c}{\bf phi4} & \multicolumn{3}{c}{\bf gemma2} & \multicolumn{3}{c}{\bf qwen2.5} & \multicolumn{3}{c}{\bf deepseek-r1}\\
        ~ & \bf F & \bf V & \bf A & \bf F & \bf V & \bf A & \bf F & \bf V & \bf A & \bf F & \bf V & \bf A & \bf F & \bf V & \bf A & \bf F & \bf V & \bf A\\
        \midrule
        none & 0.11 & 0.07 & 0.00 & 0.87 & 0.08 & 0.00 & \bf 0.97 & \bf 0.96 & 0.00 & 0.62 & 0.08 & 0.01 & 0.18 & 0.18 & 0.04 & 0.87 & 0.36 & 0.06\\
        1-small & \bf 0.92 & 0.85 & 0.00 & 0.81 & 0.60 & 0.00 & \bf 0.99 & 0.57 & 0.00 & \bf 1.00 & \bf 1.00 & \bf 0.32 & \bf 0.99 & 0.88 & 0.24 & 0.88 & 0.67 & 0.19\\
        2-small & 0.74 & 0.68 & 0.00 & \bf 0.98 & \bf 0.98 & 0.00 & 0.83 & 0.83 & 0.00 & \bf 1.00 & \bf 1.00 & 0.00 & \bf 0.96 & \bf 0.91 & 0.00 & \bf 0.93 & 0.84 & \bf 0.38\\
        1-large & \bf 0.92 & \bf 0.90 & 0.00 & \bf 0.95 & \bf 0.95 & 0.01 & \bf 0.99 & \bf 0.93 & 0.00 & \bf 1.00 & \bf 1.00 & 0.28 & \bf 0.99 & \bf 0.99 & 0.00 & 0.85 & 0.56 & 0.10\\
        2-mixed & 0.88 & 0.47 & 0.00 & \bf 1.00 & 0.73 & 0.00 & \bf 0.92 & 0.70 & 0.00 & \bf 1.00 & 0.58 & 0.15 & \bf 0.99 & \bf 0.98 & 0.02 & \bf 0.95 & 0.80 & \bf 0.36\\
        \bottomrule
    \end{tabular}
    }
\end{table*}

\subsection{Distribution of Repair Operations}

\begin{figure}[h]
    \centering
    \includegraphics[width=\columnwidth]{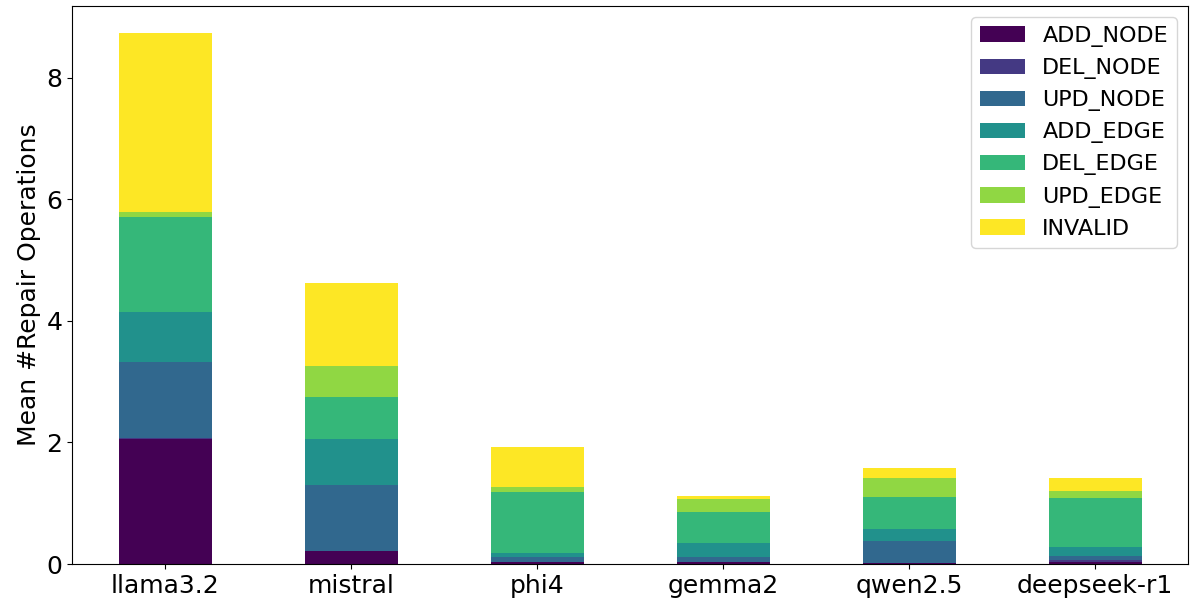}
    \caption{Distribution of Repair Operations Across Models}
    \Description{}
    \label{fig:repair_distribution}
\end{figure}

We measure the frequency of different categories of repair operations suggested.
\Cref{fig:repair_distribution} provides a visual representation of the
distribution of repair operations. We observe that \texttt{llama3.2} provides
the largest number of repair operations, also with a high percentage of
operations not adhering to the set of allowed operations. There is a positive
correlation between the number of repairs generated and number of invalid repair
operations. Thereby, models that adhere to the instruction and keep the
suggested repairs brief, result in a comparatively improved performance as
previously seen by the comparison of \Cref{fig:llm_comparison} and
\Cref{table:llm_cost}.

\subsection{Summary}

Our evaluation highlights key trade-offs between accuracy, efficiency, and
computational cost across different repair strategies. Models such as
\texttt{deepseek-r1} achieve relatively high repair validity but at the cost of
considerable increase in the response time. On the other hand, while models like
\texttt{gemma2} and \texttt{phi4} `can' balance speed and accuracy, wherein,
under specific circumstances (bold entries in \Cref{table:encoding_ablation} and
\Cref{table:example_ablation}), they are able to achieve similar accuracy scores
at much lower computational costs.

\subsection{Failure Cases and Observations}

We now examine some observed cases of failure. The majority of failures stem
from the LLMs generating an excessive number of repair operations, resulting in
more changes than necessary. An intriguing case arises in the template-based
encoding (see \Cref{table:encoding_ablation}), where models \texttt{mistral} and
\texttt{phi4} generate valid repairs for every violation but fail to produce a
single accurate repair. Specifically, \texttt{phi4} fails in $98.8\%$ of cases
(all but two) by suggesting the removal of an incorrect edge, thereby
eliminating the information that a person takes a medication. Furthermore, in
several instances, instead of proposing a single repair operation, it offers
multiple options to the user. These cases are also deemed incorrect because they
undermine the purpose of autonomous repair. Similarly, in the same setup,
\texttt{mistral} fails by recommending hallucinated updates to other nodes in
addition to the correct repairs. Such phenomenon of generating valid repairs but
drastically failing at generating correct repairs occurs in several more setups,
observable in \Cref{table:encoding_ablation} and \Cref{table:example_ablation}
by the presence of bold pair of F and V values in combination with a low A
value. To summarize, primary causes of failures are: (i)~eager generation,
(ii)~indecision, and (iii)~hallucination. There may be hope for addressing the
first two cases with stricter prompts and more training examples. On the other
hand, domain specific retrieval-augmented methods may aid in reducing
hallucinations. 
\section{Discussion and Future Work}
\label{sec:discussion}

LLMs exhibit significant sensitivity to variations in encoding methods and the
nature of examples provided in prompts. Each model responds differently, with no
consistent performance trend across architectures, highlighting the challenge of
standardizing repair strategies.

\paragraph{Response Length and Interpretability}

An architectural feature of the \texttt{deepseek-r1} models is their generation
of comprehensive prompts, explicitly including the model's internal reasoning
enclosed within \texttt{<think></think>} tags.  Furthermore, the analysis of
model outputs, as presented in \Cref{table:llm_cost}, reveals a broader trend of
increased textual generation beyond the core repair tasks. Arguably, this
mechanism provides insight into the model's decision-making process, albeit at
the cost of increased output length and processing time. The
\texttt{<think></think>} dialogues, while offering potential for enhanced
transparency, require further analysis to determine their true impact on repair
accuracy and efficiency. Future work should explore the
correlation between the explicitness of these thought processes and the model's
ability to generate valid repairs.

The following is an example of a response from \texttt{deepseek-r1} model
providing a correct repair. The model's `thought process' illustrates its
ability to understand the graph structure. The response, however, did not
utilize any real-world factual knowledge (e.g., the correctness of the
ingredient edge). This highlights another possible issue that a model may be
stumbling on a correct repair by chance.

\begin{lstlisting}
<think>
Okay, so I'm trying to figure out how to fix this
inconsistency in the graph. Let me start by
understanding what's going on here.

From the description, there's a patient (p) who
takes a medication (m). The medication has an
ingredient (i), and that same ingredient is also
an allergen for the patient. This seems
problematic because taking a medicine with an
allergen that you're allergic to could be
dangerous.

Looking at the code provided:

They have two relationships: rm connects p to m,
and rc connects m to i. Then there's another
relationship ra connecting p directly to i. The
problem is that the medication (m) contains an
ingredient (i) which the patient is allergic to,
so taking it isn't safe.

The example output suggests removing one of these
relationships or adding a new node. In this case,
maybe we should remove the rc edge because if m
has i as an ingredient, then the HAS_INGREDIENT
relationship might be incorrect in this context
since p is allergic to i.

So, by deleting the rc edge, we're removing the
connection between m and i, which could help
prevent that conflict. That makes sense because
if there's no longer a direct link from the
medication to its allergen, it avoids the
situation where the patient is taking a medication
with an allergen they're allergic to.

I don't see any need to add new nodes or edges
here; just removing rc should suffice.
</think>

<repairs>
DEL_EDGE | [rc] | -
</repairs>
\end{lstlisting}

\vspace*{-4mm}
\paragraph{Prompt Engineering and Few-Shot Learning}

Template-based encoding generally leads to better repair validity, highlighting
the importance of structured input. The current templates, formulated by humans,
demonstrate the efficacy of domain-specific knowledge in prompt design. However,
LLM-generated descriptions of inconsistencies also show promise, suggesting the
potential for automated prompt engineering. Few-shot learning, while improving
performance, can introduce inconsistencies when excessive examples are provided.
LLMs exhibit a tendency to blindly copy repairs suggested in the examples,
indicating a need for strategies to mitigate this behavior.

Our experiments reveal varied responses across different LLMs.
\texttt{deepseek-r1} benefits from 2-shot prompting, while maintaining
comparable performance in 1-shot and zero-shot settings. \texttt{qwen2.5} and
\texttt{gemma2} consistently perform well across both 1-shot and 2-shot
scenarios, exhibiting notable improvements over zero-shot. \texttt{mistral}
remains robust across all setups, with measurable gains in 2-shot
configurations. In contrast, \texttt{phi4} maintains strong performance but
experiences degradation when shifting from 1-shot to 2-shot. \texttt{llama3.2}
demonstrates substantial gains from zero-shot to 1-shot but suffers a
significant decline beyond this point. These observations underscore the
variability in LLM performance based on the number of examples provided in the
prompt, highlight the complexity of prompt tuning, and reinforce the need for
adaptive strategies that balance generalization and specificity.

\paragraph{LLM-Generated Repairs and Human Collaboration}

LLMs excel at learning structured formats--\texttt{gemma2}, \texttt{qwen2.5} and
\texttt{deepseek-r1} leading the charts--but face challenges with precision in
complex graph repairs. Ensuring syntactic correctness, mitigating
hallucinations, and optimizing token efficiency remain critical obstacles to
practical deployment. While LLM-aided graph repair demonstrates promise, a
hybrid human-in-the-loop approach could further enhance accuracy by leveraging
domain expertise to validate and refine suggested modifications.

Integrating LLMs with interactive workflows where human oversight guides
ambiguous or uncertain repairs represents a compelling future direction. This
approach would not only improve the reliability of automated repairs but also
provide valuable insights into model limitations, enabling iterative
improvements in prompt design and repair validation.

\paragraph{Explainable and Trustworthy Systems}

An additional avenue for future research lies in enhancing the explainability of
LLM-generated repairs for human users. While models can generate repair
suggestions, their justifications often remain opaque, limiting interpretability
in critical decision-making contexts. Developing mechanisms for structured
explanations, where models explicitly outline the rationale behind each repair
action, could improve trust and facilitate debugging.

\section{Limitations}
\label{sec:limitations}

This study is subject to several limitations. First, the evaluation was
conducted on a specific dataset and set of repair tasks, which may not
generalize to all graph repair scenarios. While the selected benchmarks provide
a representative testbed, future research should explore broader datasets and
real-world graph structures to ensure wider applicability.

Second, the prompt design was limited to a predefined set of templates and
examples, which may influence model performance. While our results highlight
clear trends in prompt sensitivity, further work is needed to investigate
adaptive prompt optimization techniques, such as retrieval-augmented generation
or reinforcement learning-based refinements.

Third, our computational cost analysis primarily focused on token usage and
processing time, without considering additional factors such as memory
consumption and parallelization. A more comprehensive efficiency evaluation,
particularly in large-scale deployments, could provide deeper insights into the
trade-offs between accuracy and resource constraints.

Finally, while our study evaluates LLM-generated repairs in an automated
setting, human-in-the-loop verification remains an open research direction.
Integrating active user feedback could enhance the reliability of automated
repairs and help mitigate errors in complex graph modifications. Despite these
limitations, our findings provide a strong foundation for further exploration of
LLM-assisted graph repair, underscoring the potential of these models in
structured reasoning tasks.
 
\vspace*{-1mm}

\section{Conclusions}
\label{sec:conclusion}

This work evaluates LLM-based strategies for structured graph modifications,
examining their effectiveness across different architectures and prompt
configurations. The code and results associated with the study have been made
available~\cite{llm-graph-repair}.

Our findings reveal a trade-off between accuracy, efficiency, and computational
cost. Models such as \texttt{deepseek-r1} generate highly valid repairs but
incur significant latency and token overhead, whereas lighter models like
\texttt{phi4} and \texttt{gemma2} show promise in achieving a balance between
correctness and efficiency. While LLMs demonstrate strong performance in
following a structured format, they struggle with precise graph modifications,
often generating redundant or hallucinated edits. Additionally, their
sensitivity to prompt variations underscores the necessity for adaptive tuning
strategies. Future work should, therefore, focus on domain-specific fine-tuning,
retrieval augmented generation, hybrid approaches that combine LLM-generated
repairs with rule-based heuristics, and interactive human-in-the-loop frameworks.

\vspace*{-1mm} 
\begin{acks}
The work was partially funded by the Data4Health ANR/CPJ Lyon~1 grant.
\end{acks}

\bibliographystyle{ACM-Reference-Format}
\bibliography{papers}

\end{document}